\documentclass{article} 
\usepackage[preprint]{colm2025_conference}

\usepackage{microtype}
\usepackage{hyperref}
\usepackage{url}
\usepackage{booktabs}

\usepackage{hyperref}
\usepackage{url}
\usepackage{amsmath}
\usepackage{amsthm}
\usepackage{algorithm}
\usepackage{algpseudocode}
\usepackage{multirow}
\usepackage{graphicx}
\usepackage{stfloats}
\usepackage{colortbl}
\usepackage{booktabs}
\usepackage{subcaption}
\usepackage{tcolorbox}

\usepackage{lineno}

\definecolor{darkblue}{rgb}{0, 0, 0.5}
\hypersetup{colorlinks=true, citecolor=darkblue, linkcolor=darkblue, urlcolor=darkblue}

\title{Investigating the Scaling Effect of Instruction Templates for Training Multimodal Language Model}

\author{Shijian Wang\thanks{Equal contribution} $^{~1}$, Linxin Song\footnotemark[1] $^{~2}$, Jieyu Zhang$^{3}$, Ryotaro Shimizu$^{4,5}$, Jiarui Jin$^{6}$, Ao Luo$^{8}$, \\\textbf{Yuan Lu$^{6}$, Li Yao\footnotemark[2] $^{~1}$, Cunjian Chen$^{9}$, Julian McAuley$^{4}$, Wentao Zhang$^{7}$, Hanqian Wu\thanks{Corresponding authors} $^{~1}$} \\
$^{1}$Southeast University, $^{2}$University of Southern California, $^{3}$University of Washington,\\$^{4}$University of California San Diego, $^{5}$ZOZO Research, $^{6}$Xiaohongshu Inc, \\ $^{7}$Peking University, $^{8}$Waseda University, $^{9}$Monash University
}

\newcommand\best[1]{\textcolor{red}{\textbf{#1}}}
\newcommand\secbest[1]{\textcolor{blue}{\underline{#1}}}

\newcommand{\highlight}[1]{\noindent\textbf{#1}}

\begin{document}

\ifcolmsubmission
\linenumbers
\fi

\maketitle

\begin{abstract}
 Current multimodal language model (MLM) training approaches overlook the influence of instruction templates. Previous research deals with this problem by leveraging hand-crafted or model-generated templates, failing to investigate the scaling effect of instruction templates on MLM training. In this work, we propose a programmatic instruction template generator capable of producing over 15K unique instruction templates by filling randomly sampled positional synonyms into weighted sampled meta templates, enabling us to comprehensively explore MLM's performance across various template scales in the training process. Our investigation into scaling instruction templates for MLM training demonstrates that MLM capabilities do not consistently improve with increasing template scale. Instead, optimal performance is achieved at a medium template scale. Models trained with data augmented at the optimal template scale achieve performance gains of up to 10\% over those trained on the original data and achieve the best overall performance compared with the similar-scale MLMs tuned on at most 75 times the scale of our augmented dataset. 
 The code will be publicly available at \url{https://github.com/shijian2001/TemplateScaling}. 
\end{abstract}

\vspace{-1em}
\section{Introduction}
\label{sec:intro}


Multimodal Language Models (MLMs) have revolutionized vision-language learning by performing visual instruction tuning on diverse, high-quality multimodal instruction data~\citep{liu2024visual,zhu2023minigpt, li2024llava, zhang2024provision}. However, previous studies~\citep{zhang2024task,liu2024seeing,sclar2023quantifying} reveal a critical limitation: MLMs exhibit substantial performance variability across different instruction templates. For instance, a succinct instruction and a detailed instruction can yield performance gaps exceeding 40\%~\citep{zhang2024task}. This pronounced sensitivity to instruction templates compromises the reliability of MLM evaluation and diminishes the practical utility of MLMs in downstream applications.

Recent studies have empirically demonstrated that incorporating multiple instruction templates during MLM's training process improves model performance and reduces instruction sensitivity~\citep{zhang2024provision, razavi2025benchmarking, sanh2021multitask}. However, existing approaches primarily depend on either human-designed or model-generated small-scale templates, which suffer from limitations such as high costs, inherent design biases, and limited diversity in instruction formulations.
Considering the success of scaling up training data significantly improves model's performance~\citep{fan2024scaling, kaplan2020scaling} and the fact that multi-template training can improve MLM, this raises a critical question: \textit{How many instruction templates should be used during training to optimize MLM performance}?


\begin{figure}[t]
  \includegraphics[width=\textwidth]{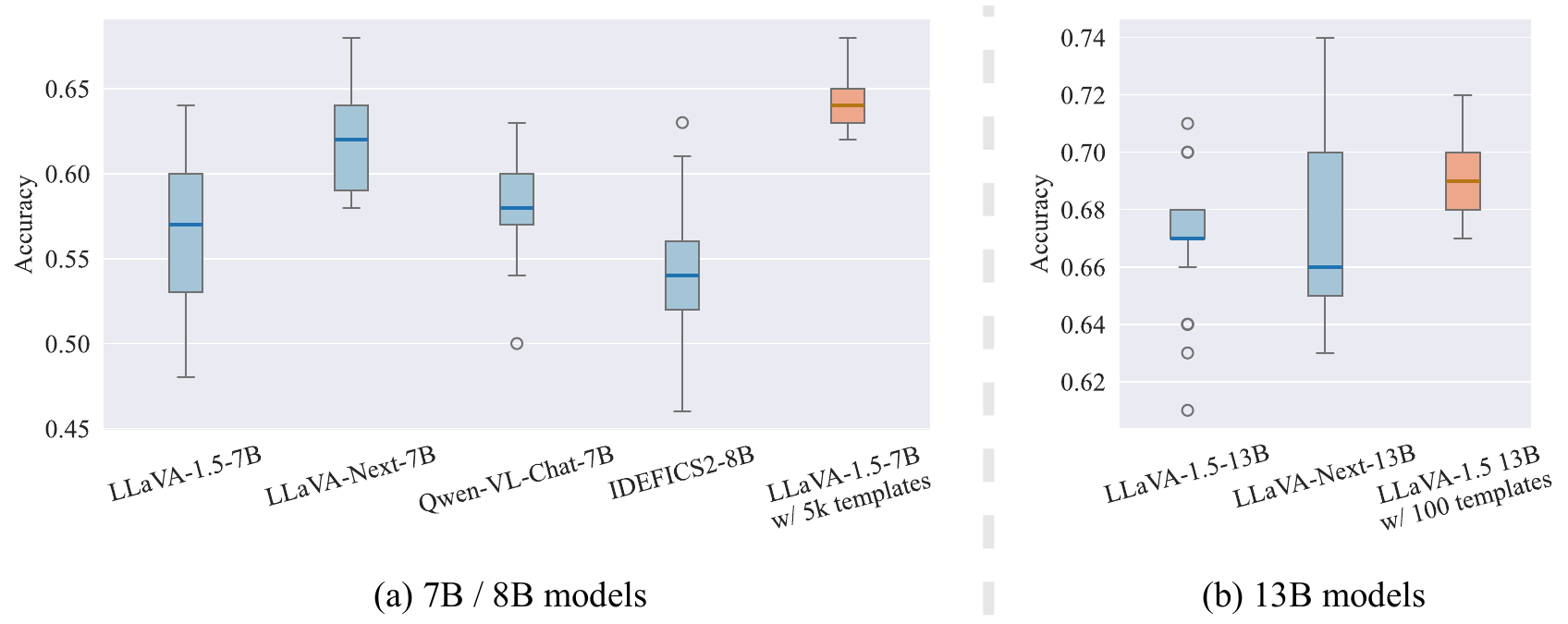}
  \caption{\textbf{Training with the optimal template scale significantly improves MLM's performance and reduces the performance variance.} LLaVA-1.5-7B trained with 5K templates and LLaVA-1.5-13B trained with 100 templates achieve the highest average performance and the lowest performance variance among similar-scale MLMs on the SeedBench dataset, evaluated across 25 held-out instruction templates that are not included in training.}
  \label{fig:banner}
\end{figure}

To systematically investigate the scaling effect of instruction templates for MLM training, we propose a \textit{\textbf{programmatic instruction template generator}} that leverages diverse meta templates to produce semantically equivalent instruction templates automatically and scalably. 
Our template generator can construct diverse instruction templates by random sampling from carefully curated word and phrase spaces to populate predefined placeholders, enabling the efficient generation of semantically consistent yet diverse instruction templates at scale. 
Our method can produce an extensive template space comprising 15K visual instruction templates.
To ensure the diversity of sampled instruction templates from our template generator, we use a sentence-pattern tree organizational framework based on grammatical structures complemented by an efficient diverse sampling algorithm. This programmatic approach ensures the generation of instruction templates that maximize diversity across multiple dimensions, including grammatical construction, lexical choice, and symbolic representation. 


Leveraging our programmatic template generator, we finetune two widely-used MLMs (LLaVA-1.5-7B and LLaVA-1.5-13B)~\citep{liu2024improved} and conduct a series of experiments by performing visual instruction tuning on the same dataset while varying the scale of instruction templates (from 10 to 15K).
Our study reveals that the performance of MLMs does not consistently improve with the increasing scale of instruction templates. Instead, MLMs achieve the best general capabilities at a medium template scale, which varies with the model's parameter size. We find LLaVA-1.5-7B's performance peaks at 5K templates and LLaVA-1.5-13B peaks at 100 templates.
We further compare our models trained under the optimal template scale with other MLMs fine-tuned on a significantly larger scale—up to 75.19 times the size of our instruction tuning datasets. 
Evaluation across five benchmarks reveals that our tuned models achieve the best overall performance (We showcase the comparison results on the SeedBench~\citep{li2023seed} dataset in Figure~\ref{fig:banner}), thereby demonstrating the capacity of training with appropriate template scale to enhance MLMs in a data-efficient and cost-effective manner. 
Additionally, our analysis reveals that, compared to the original model, fine-tuning with the optimal template scale results in a substantial reduction in performance variance across various out-of-domain instruction templates. Our approach not only confirms the practical utility of the scaling effect of instruction templates but also provides promising insights into efficient strategies for improving MLMs.
We summarize our main contributions as follows.

(1) We introduce a novel programmatic instruction template generator that enables fast and scalable generation of diverse, semantically equivalent instruction templates.

(2) We comprehensively investigate the scaling effect of instruction templates for MLM training, demonstrating that MLM capabilities do not monotonically improve with increasing template scale and instead peak at a medium template scale.

(3) We propose a simple yet effective approach to enhance visual instruction tuning by augmenting the original instruction tuning dataset with the optimal scale of templates we investigated. Our extensive experiments demonstrate its effectiveness.

\section{Programmatically Scaling Instruction Templates}
\label{sec:method}

\begin{figure}[t]
    \centering
    \includegraphics[width=\linewidth]{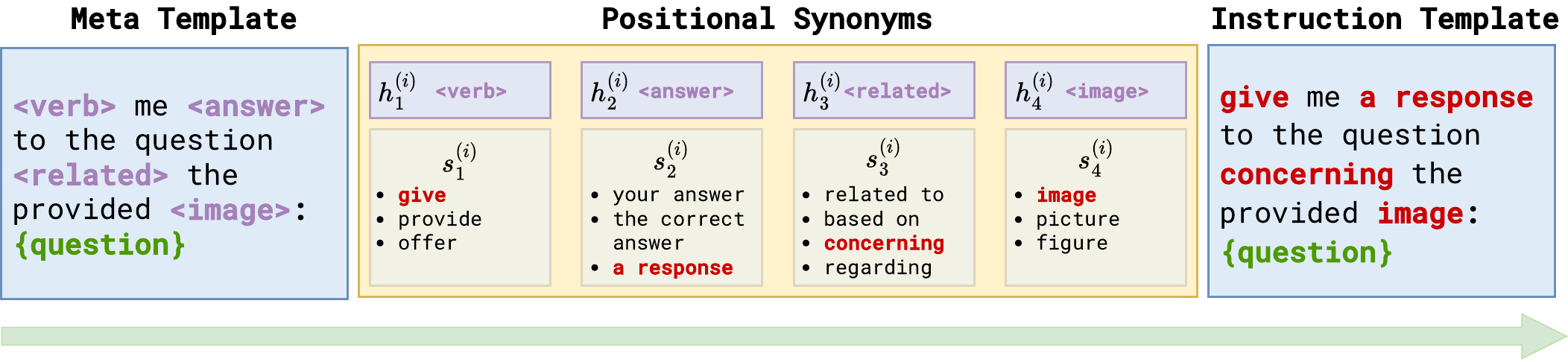}
    \caption{Example of the instruction template generation through a meta template.}
    \label{fig:method}
\end{figure}

To investigate the scaling effect of instruction templates in MLM's visual instruction tuning, we propose a programmatic instruction template generator. Our template generator can efficiently produce diverse, grammatically correct, and semantically consistent instruction templates. Specifically, we generate instruction templates by programmatically filling the pre-defined placeholders in a \textit{meta template} with randomly sampled positional synonyms (phrases), ensuring flexibility and diversity while keeping the original meaning (Sec.~\ref{sec:meta-templates}). We organize our meta templates in a \textit{sentence pattern tree}, along with a diverse template sampling algorithm to ensure the sampling probability across all instruction templates is uniformly distributed (Sec.~\ref{sec:pattern-tree}).

\subsection{Meta Templates}
\label{sec:meta-templates}
We design meta template $p_i, i\in\{1,...,N\}$ as a formal blueprint for constructing instruction templates, consisting of a sequence of fixed string segments interspersed with placeholder $\langle h^{(i)}_j \rangle, j\in\{1,...,M_i\}$, where $M_i$ is the number of placeholders. We associate each placeholder $\langle h^{(i)}_j \rangle$ with a predefined set of synonyms (phrases) $s_j^{(i)}$. We design $s_j^{(i)}$ according to the semantic position of $\langle h^{(i)}_j \rangle$, including nouns, verbs, adjectives, or more abstract functional tokens pertinent to the context of the instruction. The potential template variations $\mathcal{T}(p_i)$ grow combinatorially as $\mathcal{T}(p_i) = \prod_{j=1}^{M_i} |s_j^{(i)}|$, where $|s_j^{(i)}|$ is the size of each synonym set.
Fixed strings establish the foundational sentence structure, ensuring grammatical correctness and semantic coherence, while placeholders introduce flexibility and diversity, enabling the rapid generation of varied, high-quality instruction templates. We present the example of instruction template generation through a meta template in Figure~\ref{fig:method}.
To ensure the diversity of generated visual instruction templates, we design 24 meta templates, yielding a template space capable of producing 15K distinct instruction templates.

\subsection{Diverse Template Sampling}
\label{sec:pattern-tree}
\highlight{Sentence pattern tree.} 
We build a sentence pattern tree to systematically organize our meta templates.
We use $T = (V, E)$ to denote the sentence pattern tree, where $V$ is the set of sentence patterns and $E$ is the edge between related sentence patterns. $T$ consists of four levels, ranging from coarse-grained to fine-grained, according to the taxonomy of sentence patterns. We use level 1 to represent the highest level of a sentence pattern, including declarative and imperative sentences. Level 2 decomposes Level 1 into simple, complex, and compound sentences. Level 3 further breaks Level 2 into subject-predicate, subject-predicate-object, subject-subject, noun clause, gerund clause, and linking clauses. Leaves in the final Level 4 represent the meta templates belonging to the above parent nodes. 
Building on the tree framework, we can perform weighted sampling on Level 4 according to vertex features from Level 1 to Level 3.
We present the details of our sentence pattern trees with diverse meta templates in Appendix~\ref{sec:generator}.

\highlight{Weighted sampling through sentence pattern tree.}
To achieve diverse sampling across the extensive template space, we implement a top-down weighted sampling approach within the sentence pattern tree. Specifically, our approach begins by assigning a weight to each tree node. The weight of each leaf node $\ell^{(i)}$ corresponds to the number of potential templates that can be generated by the associated meta template $p_i$. These weights accumulate progressively up each level of the tree. The weight $w_v$ of each node $v \in V$ at any level represents the sum of weights of its descendant nodes in the next level. The detailed procedure for weight accumulation is outlined in Algorithm~\ref{alg:weight-accumulation}.
During the template sampling process, we select nodes in a top-down manner, with the probability of sampling each node $v$ at a given level proportional to $w_v$. Upon reaching a leaf node corresponding to a meta template, we programmatically fill the placeholders in the meta template with randomly selected positional synonyms.
This process ensures that the sampling probability across all instruction templates remains uniform, promoting diversity in generated templates while preserving the semantic consistency of each instruction template.
We describe details of the weighted sampling algorithm in Algorithm~\ref{alg:template-generation}. 

\begin{algorithm}[t]
\caption{Weight Accumulation}
\label{alg:weight-accumulation}
\begin{algorithmic}[1]
\Procedure{AccumulateWeights}{$T$}
    \For{each leaf node $v$ in $T$}
        \State $w(v) \gets \text{NumTemplates}(v)$ \Comment{Set weight to number of potential generated templates in the leaf}
    \EndFor
    \For{each non-leaf node $v$ in $T$ in reverse topological order}
        \State $C \gets \text{children}(v)$ \Comment{Retrieve children of $v$}
        \State $w(v) \gets \sum_{c \in C} w(c)$ \Comment{Sum the weights of child nodes}
    \EndFor
    \State \textbf{return} $T$ \Comment{Return tree with accumulated weights}
\EndProcedure
\end{algorithmic}
\end{algorithm}

\begin{algorithm}[t]
\caption{Weighted Sampling and Template Generation}
\label{alg:template-generation}
\begin{algorithmic}[1]
\Procedure{GenerateTemplate}{$T$}
    \State $v \gets v_0$ \Comment{Initialize at the root node of $T$}
    \While{$v$ is not a leaf node}
        \State $C \gets \text{children}(v)$ \Comment{Retrieve child nodes of $v$}
        \State $W \gets \{w(c) : c \in C\}$ \Comment{Collect weights of child nodes}
        \State $v \gets \text{WeightedRandomChoice}(C, W)$ \Comment{Select a child node based on weights}
    \EndWhile
    \State $p \gets \text{pattern}(v)$ \Comment{Retrieve the meta template from the selected leaf node}
    \For{each placeholder $\langle h_j \rangle$ in $p$}
        \State $S_j \gets \text{synonyms}(\langle h_j \rangle)$ \Comment{Retrieve synonyms for the placeholder}
        \State $s_j \gets \text{UniformRandomChoice}(S_j)$ \Comment{Randomly select a synonym}
        \State Replace $\langle h_j \rangle$ in $p$ with $s_j$ \Comment{Substitute placeholder with synonym}
    \EndFor
    \State \textbf{return} $p$ \Comment{Return the constructed instruction template}
\EndProcedure
\end{algorithmic}
\end{algorithm}

\section{Investigating Scaling Instruction Templates on MLM Training}
\label{sec:vsft-scaling}
To investigate the scaling effect of instruction templates in MLM's visual instruction tuning, we train multiple model variants using the same instruction tuning dataset while varying the scale of instruction templates. We then evaluate these template-tuned models across various benchmark datasets to observe the impact of the instruction template scale on MLM performance. We first present our experimental setup (Sec~\ref{sec:exp-setup}), followed by the experimental results and analysis (Sec~\ref{sec:scaling-exp}).

\begin{figure*}[t!]
    \centering

    \begin{subfigure}{\textwidth}
      \begin{subfigure}{0.19\textwidth}
        \centering
        \caption*{BLINK}
        \includegraphics[width=\linewidth]{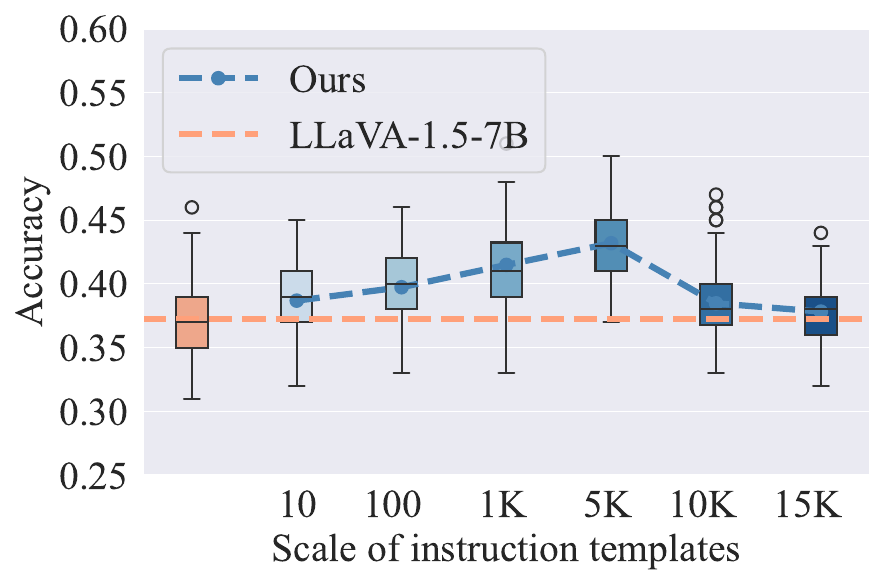}
      \end{subfigure}\hfill
      \begin{subfigure}{0.19\textwidth}
        \centering
        \caption*{MMBench}
        \includegraphics[width=\linewidth]{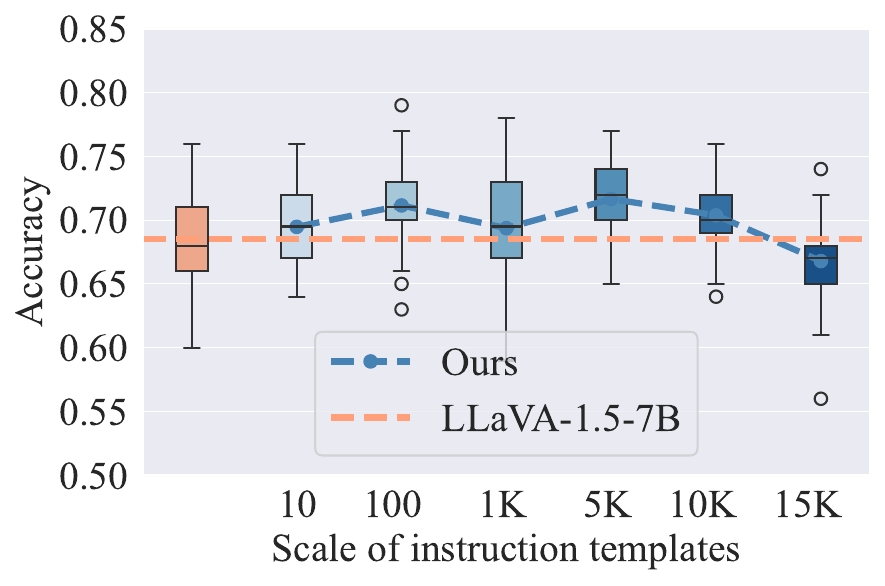}
      \end{subfigure}\hfill
      \begin{subfigure}{0.19\textwidth}
        \centering
        \caption*{SeedBench}
        \includegraphics[width=\linewidth]{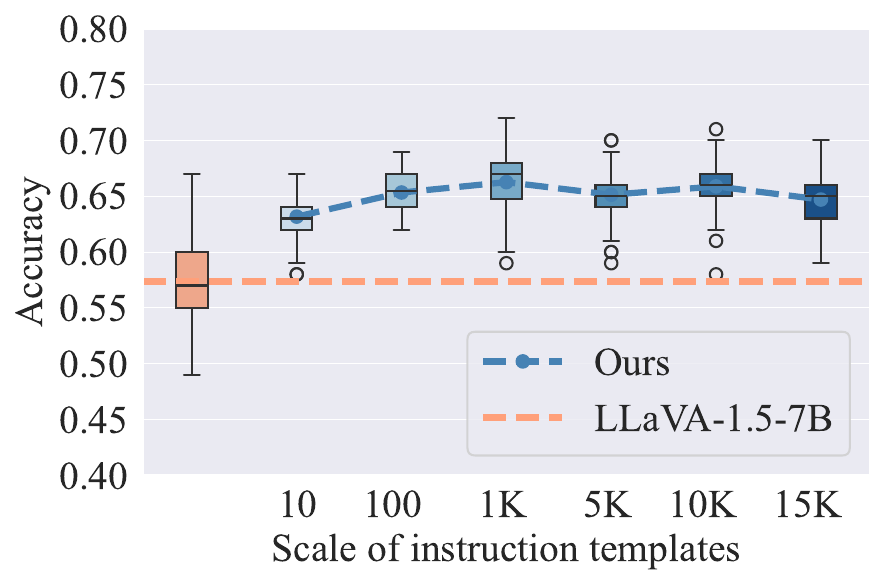}
      \end{subfigure}\hfill
      \begin{subfigure}{0.19\textwidth}
        \centering
        \caption*{TMA}
        \includegraphics[width=\linewidth]{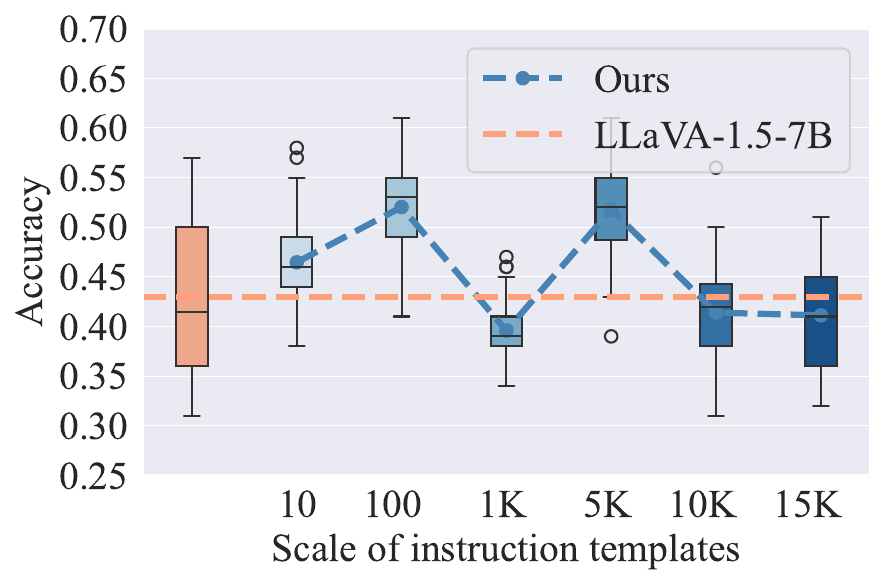}
      \end{subfigure}\hfill
      \begin{subfigure}{0.19\textwidth}
        \centering
        \caption*{MMMU}
        \includegraphics[width=\linewidth]{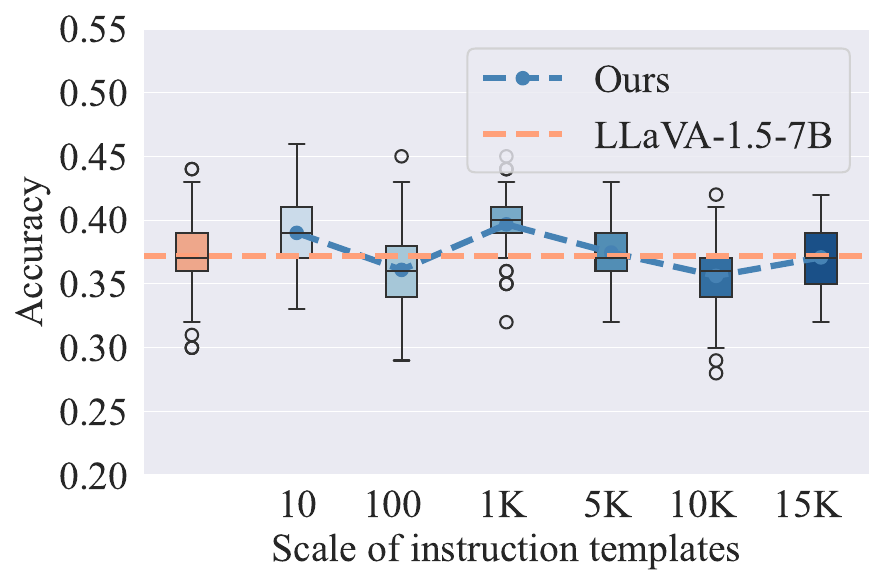}
      \end{subfigure}
    \caption{Evaluation of 7B models on in-domain templates.}
    \end{subfigure}

    \begin{subfigure}{\textwidth}
      \begin{subfigure}{0.19\textwidth}
        \centering
        \caption*{BLINK}
        \includegraphics[width=\linewidth]{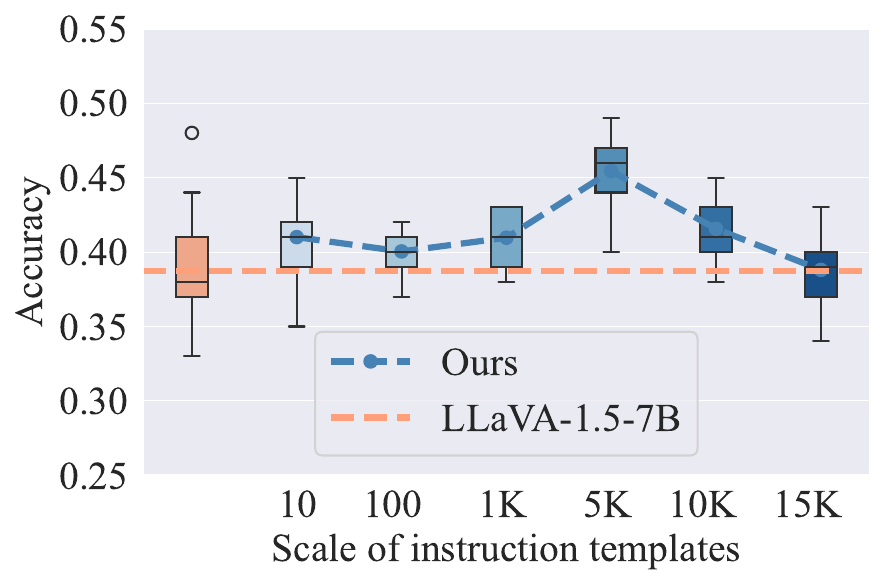}
      \end{subfigure}\hfill
      \begin{subfigure}{0.19\textwidth}
        \centering
        \caption*{MMBench}
        \includegraphics[width=\linewidth]{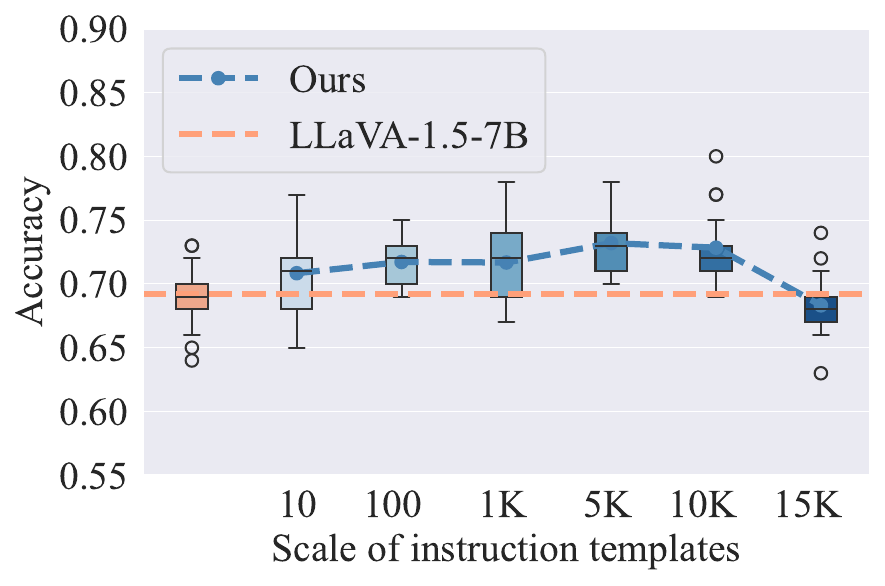}
      \end{subfigure}\hfill
      \begin{subfigure}{0.19\textwidth}
        \centering
        \caption*{SeedBench}
        \includegraphics[width=\linewidth]{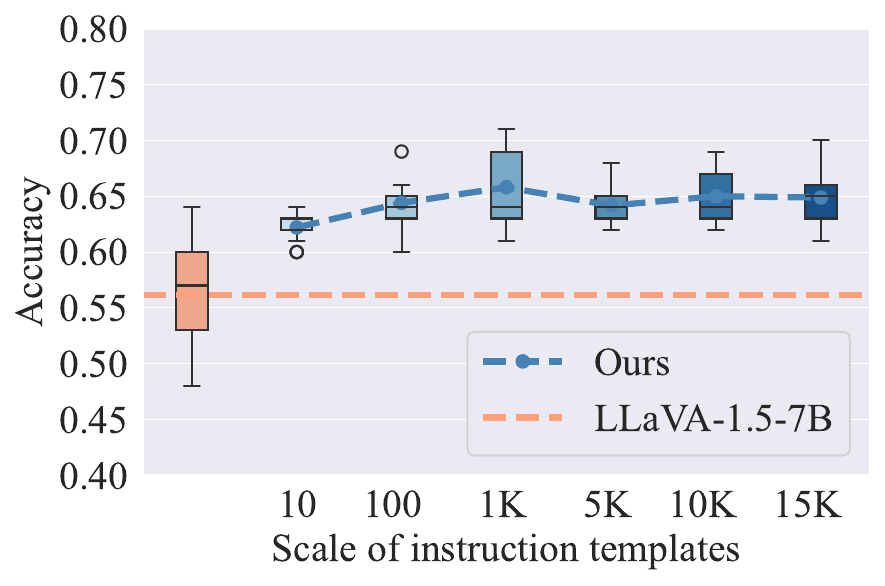}
      \end{subfigure}\hfill
      \begin{subfigure}{0.19\textwidth}
        \centering
        \caption*{TMA}
        \includegraphics[width=\linewidth]{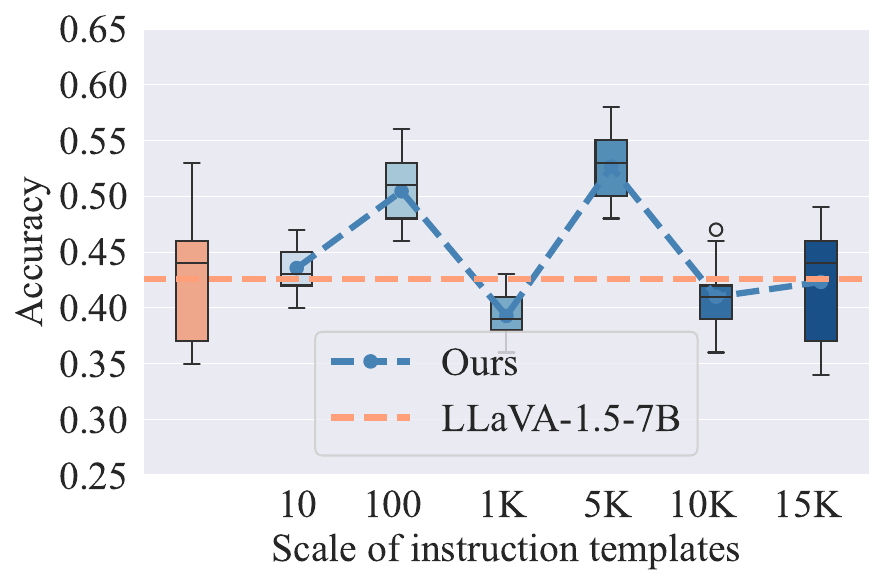}
      \end{subfigure}\hfill
      \begin{subfigure}{0.19\textwidth}
        \centering
        \caption*{MMMU}
        \includegraphics[width=\linewidth]{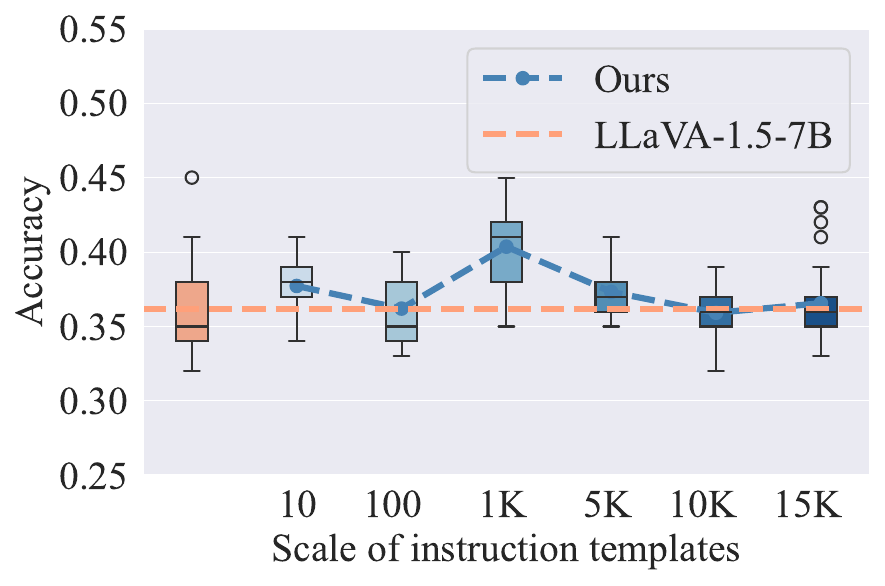}
      \end{subfigure}
    \caption{Evaluation of 7B models on out-of-domain templates.}
    \end{subfigure}

    \begin{subfigure}{\textwidth}
      \begin{subfigure}{0.19\textwidth}
        \centering
        \caption*{BLINK}
        \includegraphics[width=\linewidth]{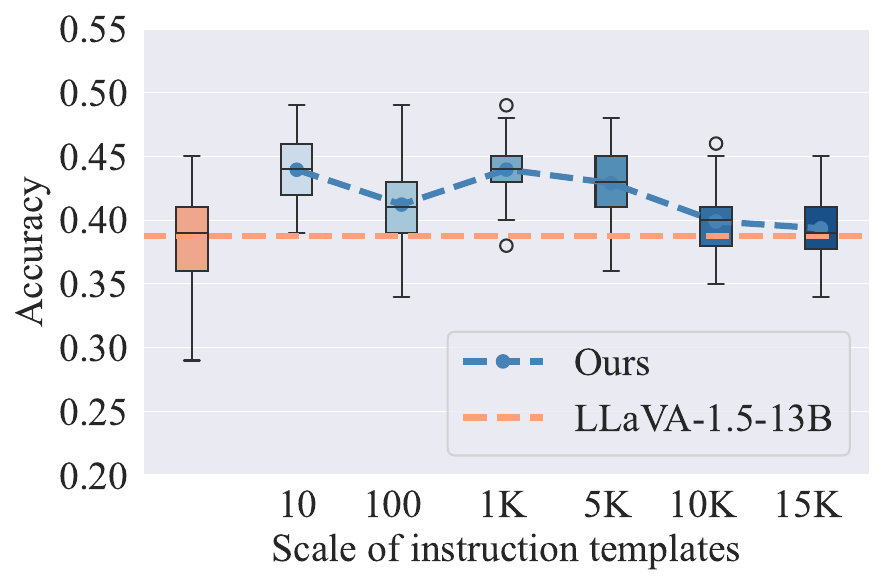}
      \end{subfigure}\hfill
      \begin{subfigure}{0.19\textwidth}
        \centering
        \caption*{MMBench}
        \includegraphics[width=\linewidth]{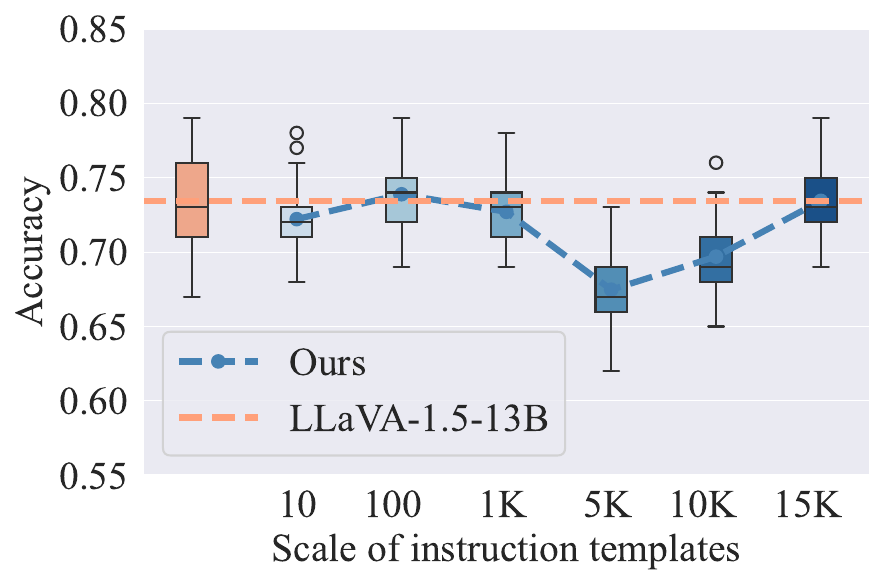}
      \end{subfigure}\hfill
      \begin{subfigure}{0.19\textwidth}
        \centering
        \caption*{SeedBench}
        \includegraphics[width=\linewidth]{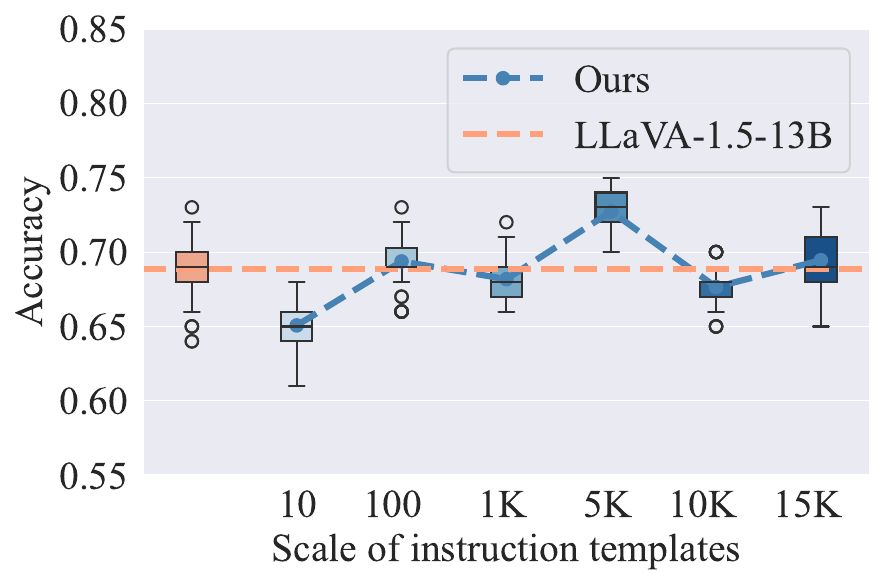}
      \end{subfigure}\hfill
      \begin{subfigure}{0.19\textwidth}
        \centering
        \caption*{TMA}
        \includegraphics[width=\linewidth]{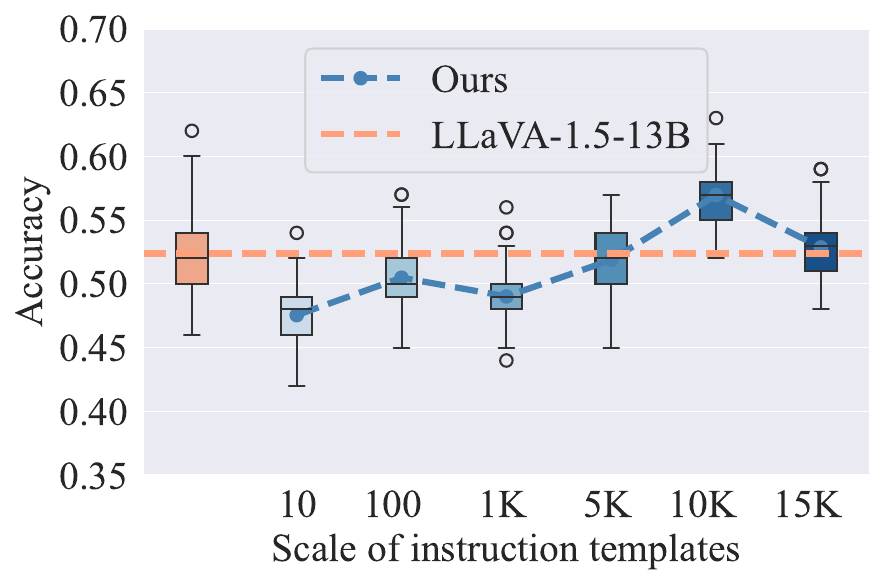}
      \end{subfigure}\hfill
      \begin{subfigure}{0.19\textwidth}
        \centering
        \caption*{MMMU}
        \includegraphics[width=\linewidth]{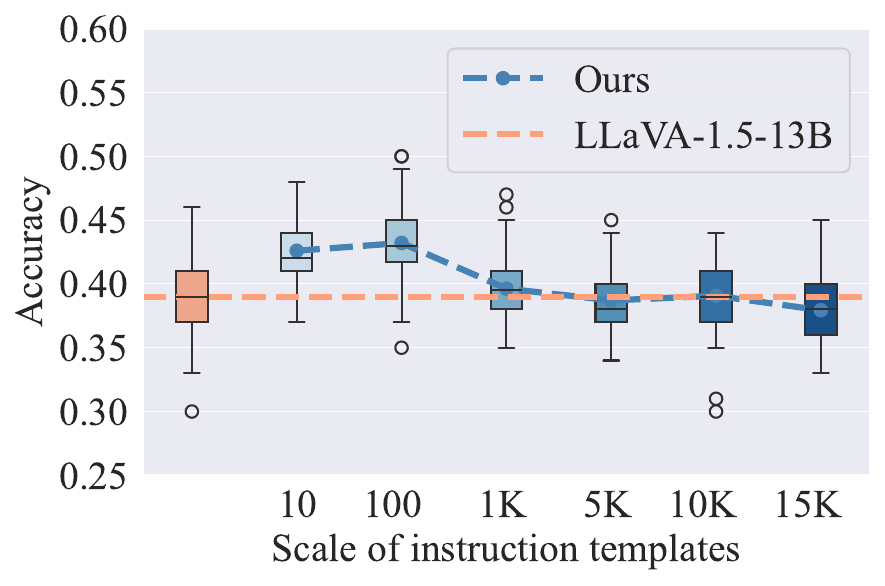}
      \end{subfigure}
    \caption{Evaluation of 13B models on in-domain templates.}
    \end{subfigure}

  \begin{subfigure}{\textwidth}
      \begin{subfigure}{0.19\textwidth}
        \centering
        \caption*{BLINK}
        \includegraphics[width=\linewidth]{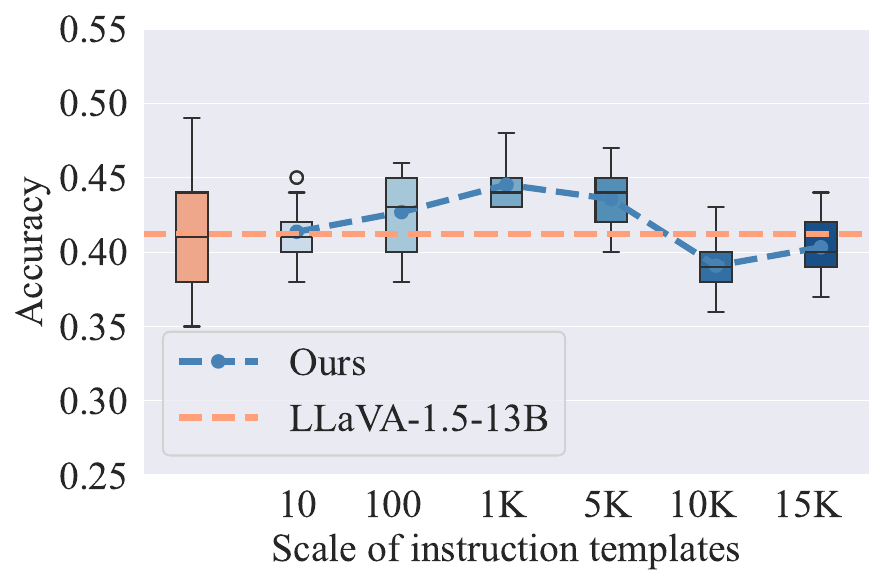}
      \end{subfigure}\hfill
      \begin{subfigure}{0.19\textwidth}
        \centering
        \caption*{MMBench}
        \includegraphics[width=\linewidth]{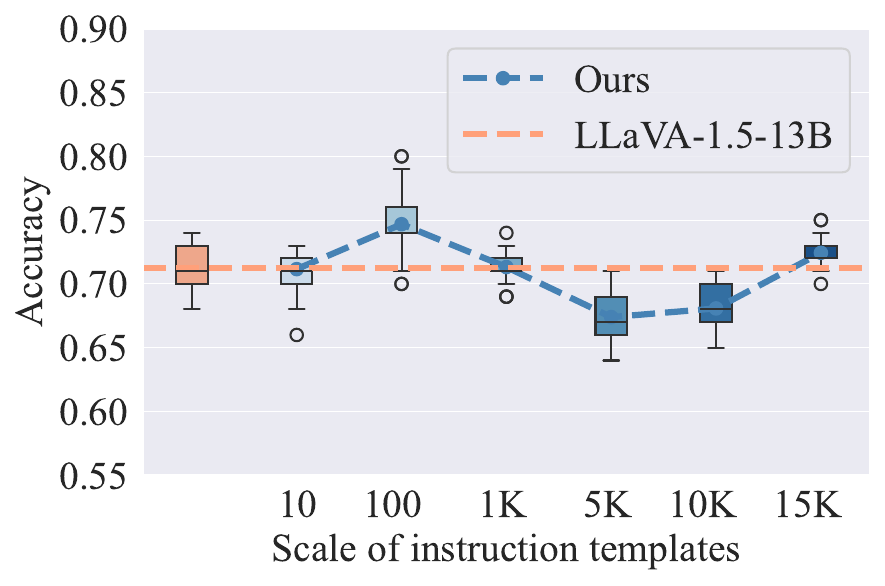}
      \end{subfigure}\hfill
      \begin{subfigure}{0.19\textwidth}
        \centering
        \caption*{SeedBench}
        \includegraphics[width=\linewidth]{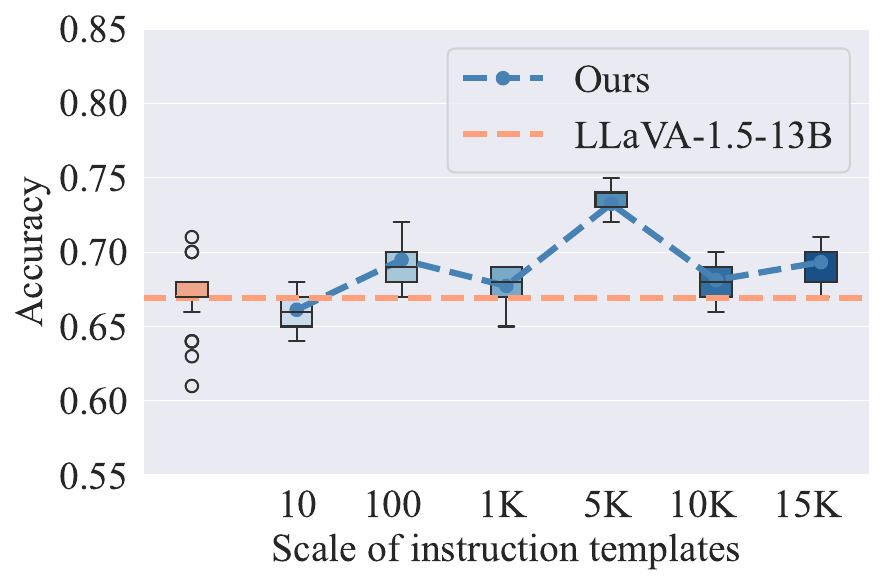}
      \end{subfigure}\hfill
      \begin{subfigure}{0.19\textwidth}
        \centering
        \caption*{TMA}
        \includegraphics[width=\linewidth]{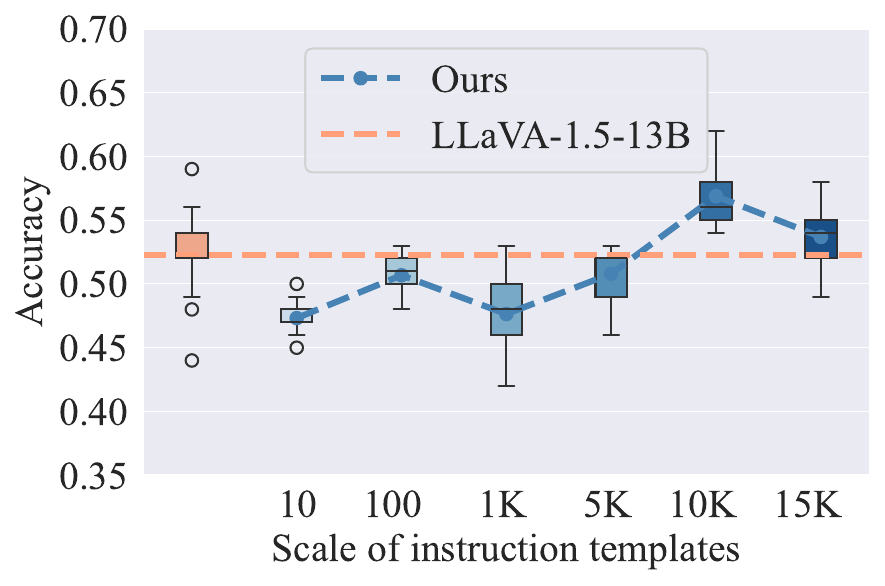}
      \end{subfigure}\hfill
      \begin{subfigure}{0.19\textwidth}
        \centering
        \caption*{MMMU}
        \includegraphics[width=\linewidth]{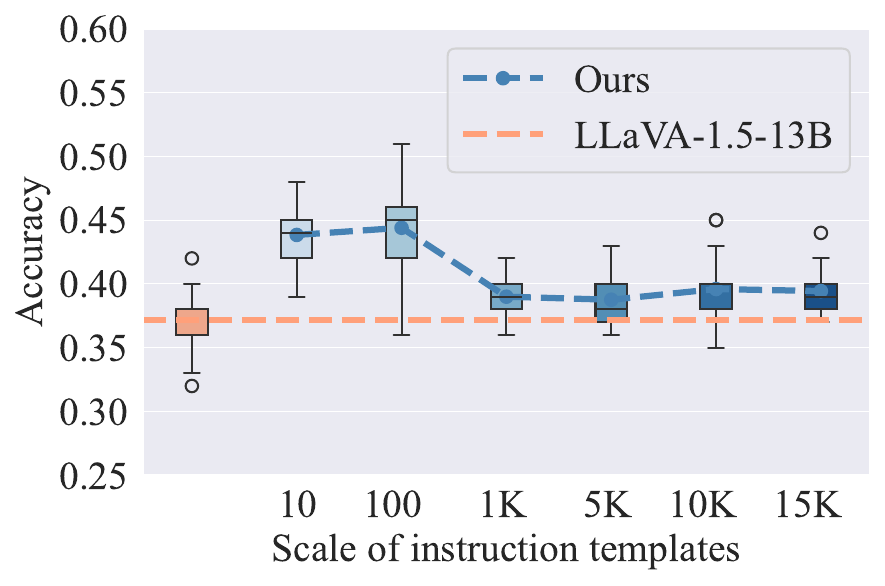}
      \end{subfigure}
    \caption{Evaluation of 13B models on out-of-domain templates.}
  \end{subfigure}
  
  \caption{Scaling trends of MLM performance with increasing template scale on each benchmark dataset. We also show the performance spread across models and datasets. \textbf{Optimal template scale vary across different datasets.}}
  \label{fig:scaling-dataset}
\end{figure*}

\subsection{Experiment Setup}
\label{sec:exp-setup}

\highlight{Training configurations.} 
We trained template-tuned models based on the pretrained checkpoints: LLaVA-1.5-7B-Base and LLaVA-1.5-13B-Base, which are strong starting points for visual instruction tuning due to the open-source nature of data and models in this series.
We used Low-Rank Adaptation (LoRA)~\citep{hu2021lora} to train all models under the same hyperparameter settings. We used a batch size of 128 and a learning rate of $2\times10^{-5}$ with a cosine decay schedule. The learning rate warmup ratio is set to 0.03. We used the AdamW~\citep{AdamW} optimizer and performed fine-tuning with DeepSpeed\footnote{\url{https://github.com/microsoft/DeepSpeed}} at stage 3.
We trained all models with 16 $\times$ A100 (40G).

\begin{figure*}[ht]
  \centering
  \begin{subfigure}{0.24\textwidth}
    \centering
    \includegraphics[width=\linewidth]{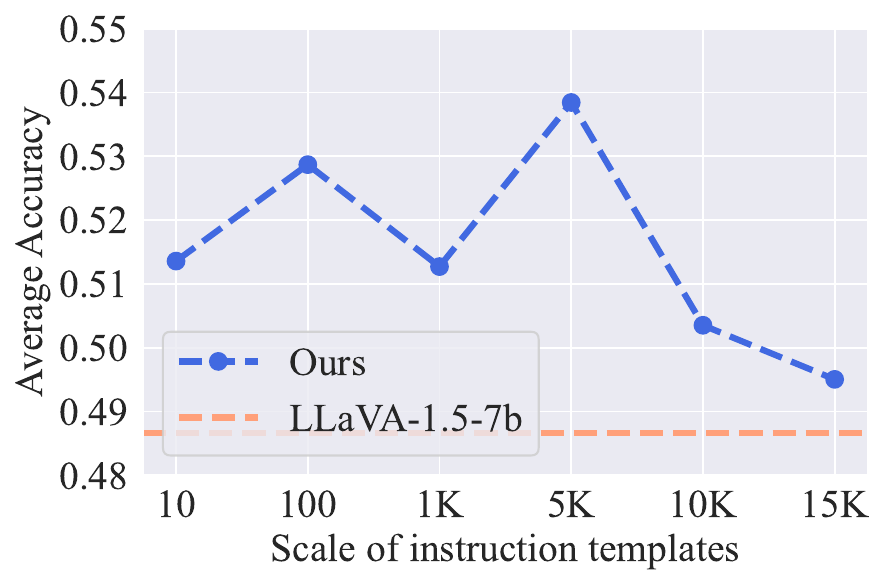}
    \caption{7B models on in-domain templates.}
  \end{subfigure}\hfill
  \begin{subfigure}{0.24\textwidth}
    \centering
    \includegraphics[width=\linewidth]{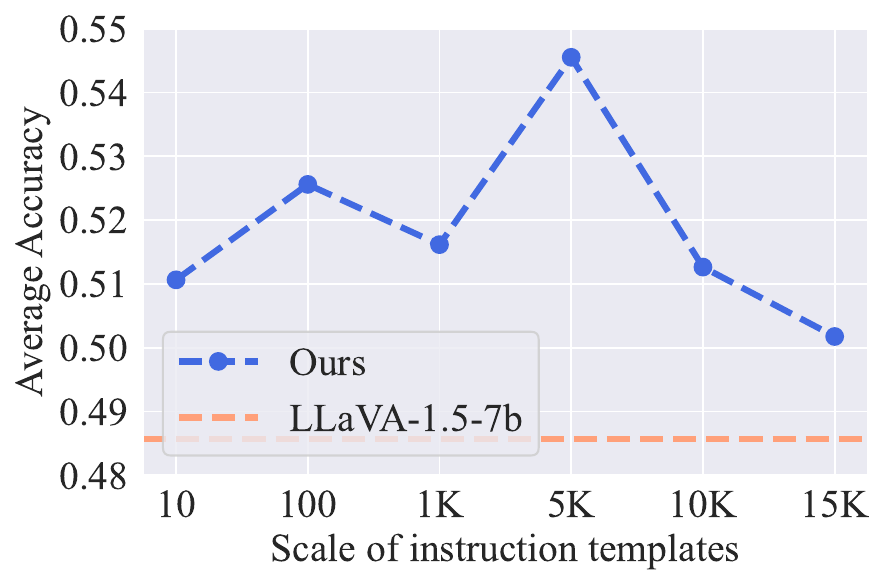}
    \caption{7B models on out-of-domain templates.}
  \end{subfigure}\hfill
  \begin{subfigure}{0.24\textwidth}
    \centering
    \includegraphics[width=\linewidth]{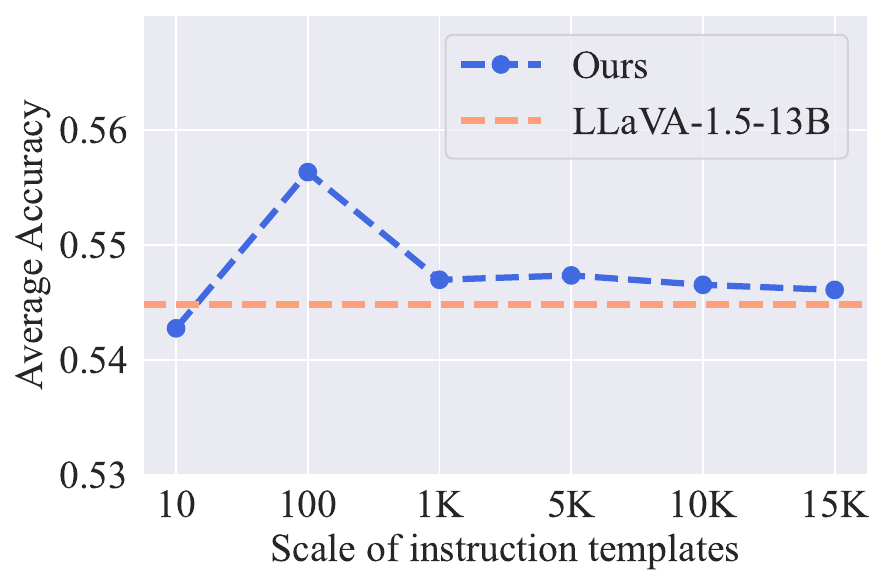}
    \caption{13B models on in-domain templates.}
  \end{subfigure}\hfill
  \begin{subfigure}{0.24\textwidth}
    \centering
    \includegraphics[width=\linewidth]{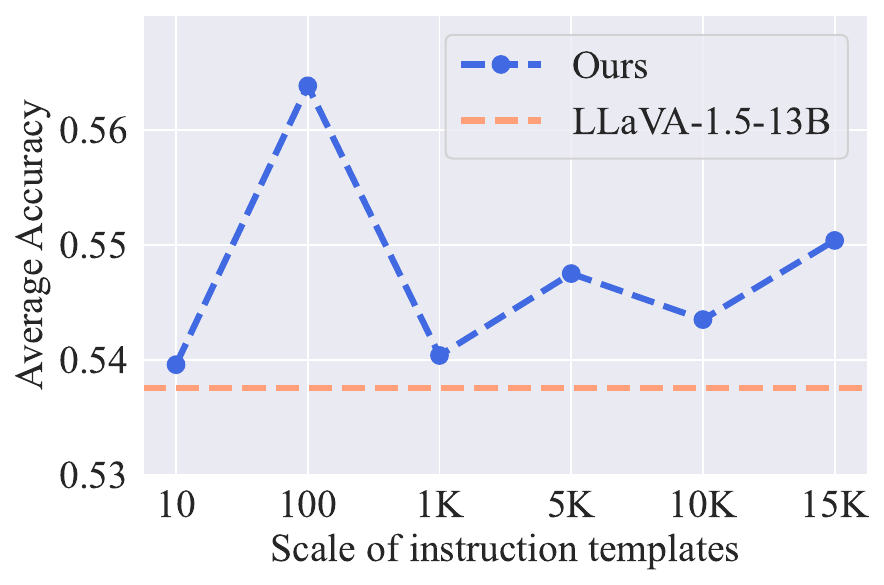}
    \caption{13B models on out-of-domain templates.}
  \end{subfigure}\hfill
  
  \caption{Scaling trend of MLM performance with increasing template scale on the average performance across five benchmarks. \textbf{There exists an optimal template scale for MLM's general capabilities, with stronger models requiring a smaller template scale.}}
  \label{fig:scaling-avg}
\end{figure*}


\highlight{Scaling instruction templates in training data.}
We constructed six template-augmented versions of the original 665K-scale multimodal instruction-following data\footnote{\url{https://huggingface.co/datasets/liuhaotian/LLaVA-Instruct-150K/blob/main/llava_v1_5_mix665K.json}} (provided by the LLaVA-1.5 series) by applying randomly sampled 10, 100, 1K, 5K, 10K, and 15K templates from our programmatic template generator.  Without introducing additional data sources, we applied instruction templates to the instruction part of the training data, resulting in template-diversified training datasets that maintain the same size as the original. 
The enhanced datasets were subsequently used to finetune the pretrained LLaVA-1.5-7B-Base and LLaVA-1.5-13B-Base models. We trained a total of \textbf{twelve} models, comprising six models with 7B parameters and six models with 13B parameters.

\highlight{Benchmark datasets.}
To comprehensively examine the performance of our tuned models trained with different template scales across diverse tasks and domains, we conduct the evaluation using five popular Visual Question Answering (VQA) benchmark datasets: BLINK~\citep{fu2024blink}, SeedBench~\citep{li2023seed}, MMBench~\citep{liu2025mmbench}, TaskMeAnything~\citep{zhang2024task}, and MMMU~\citep{yue2024mmmu}.  
Each data point in the above datasets contains an image or multiple images, a question, several choices, and a correct answer. We filter these datasets to retain only the single-image samples for our evaluation.
Specifically, we randomly select 100 data points for each dataset according to their category distribution, then combine each data point with instruction templates to test.
To evaluate the robustness of these template-tuned models, we conducted evaluations under the following two evaluation template settings.

\highlight{(1) In-domain templates:} We generated 100 templates using our template generator, which our template-tuned models have encountered during training. 

\highlight{(2) Out-of-domain templates:} To assess the generalization ability of these template-tuned models, we manually wrote 25 templates that are outside the template space of our template generator. These templates serve as a held-out set for evaluation.


Populating evaluation data with the two template sets yields two new templated benchmark datasets with 10K and 2.5K samples for each original dataset.

\highlight{Evaluation Protocol.}
We fix the choice order according to the original dataset to eliminate this confounder and focus solely on the effects of template scale on model performance~\citep{zheng2023large}.
To retrieve answers from MLMs' replies, we follow~\citep{zhang2024task} and adopt a two-step approach. First, we apply a string-matching algorithm to determine if the model's output matches any of three specific option representations: (1) the option identifier, e.g., \textit{(A)}; (2) the option content, e.g., \textit{cat}; or (3) both the identifier and the name, e.g., \textit{(A) cat}. If no direct match is identified, we employ a sentence-transformer~\citep{reimers2019sentence} to calculate the embedding similarity between the model's output and each answer option, selecting the option with the highest similarity as the predicted answer.
We adopt the answer accuracy on each dataset as our evaluation metric.

\subsection{Comparing MLMs on Different Template Scales}
\label{sec:scaling-exp}
Figure~\ref{fig:scaling-dataset} provides detailed scaling curves of MLM performance with increasing template scale on each individual benchmark, while Figure~\ref{fig:scaling-avg} illustrates the scaling curves of the average performance across all datasets with increasing template scale. 
These results reveal three main findings.

\highlight{Training with diverse templates can improve MLMs.}
As illustrated in Figure~\ref{fig:scaling-dataset} and Figure~\ref{fig:scaling-avg}, models trained with diverse instruction templates, ranging from 10 to 15K templates, outperform those trained solely on the original instruction tuning data in most cases. This improvement is evident in the average performance and individual performance across all five benchmark datasets and holds true across different model parameter sizes (7B and 13B) as well as for both in-domain and out-of-domain evaluation template settings. These results underscore the significance of our investigation into the scaling effects for MLM training.

\highlight{Optimal template scale vary across datasets.}
As shown in Figure~\ref{fig:scaling-dataset}, the scaling trend of MLM performance with increasing template scale exhibits significant variability across different datasets, with the optimal template scale differing for each dataset. Furthermore, we observed that an inappropriate template scale can lead to a decrease in performance or an increase in performance fluctuation range compared to the original model on certain datasets, highlighting the significance of finding the optimal template scale.

\highlight{MLM capability presents clear scaling trend with increasing template scale.}
As illustrated in Figure~\ref{fig:scaling-avg}, the model's average performance across all five datasets exhibits a consistent scaling trend, initially increasing before declining, with peak performance achieved at a medium template scale. This trend holds across different model sizes (7B and 13B parameters) and evaluation settings (in-domain and out-of-domain templates). However, the optimal template scale varies depending on model capacity: the 7B model reaches peak performance at 5K templates, whereas the 13B model achieves its best results at a significantly smaller scale of 100 templates. This discrepancy suggests that models with stronger baseline capabilities (e.g., the 13B model) require fewer templates to attain optimal performance. Furthermore, while Figure~\ref{fig:scaling-dataset} demonstrates that model performance exhibits dataset-specific variability at smaller template scales, the performance consistently declines as the template scale increases beyond a certain threshold, demonstrating that the optimal template scale lies within a medium range, eliminating the need for exhaustive large-scale searches.

\section{Visual Instruction Tuning on the Optimal Template Scale}
\label{sec:template-tuning}

To demonstrate the practical impact of the scaling effect of instruction templates for training MLMs, we compare the performance of our tuned models trained on the optimal template scale against other prominent MLMs of similar parameter sizes. We first outline the experimental setup (Sec.~\ref{sec:vsft-main-setup}), then detail the comparison results and analysis (Sec.~\ref{sec:vsft-main-exp}). 

\subsection{Experiment Setup}
\label{sec:vsft-main-setup}

\highlight{Our method.}
We selected our best-performing template-tuned models—LLaVA-1.5-7B trained with 5K templates, and LLaVA-1.5-13B trained with 100 templates—to compare against other prominent MLMs of comparable scales. 

\highlight{Baselines.}
To establish our baseline models, we used original instruction data to perform conventional visual instruction tuning on the LLaVA-1.5-7B-Base and LLaVA-1.5-13B-Base, yielding LLaVA-1.5-7B and LLaVA-1.5-13B~\citep{liu2024improved}, which serve as our primary baselines. In addition, for the 7B parameter size, we selected LLaVA-Next-7B~\citep{liu2024llavanext}, Qwen-VL-7B and Qwen-VL-Chat-7B~\citep{bai2023qwen}, and IDEFICS2-8B~\citep{laurenccon2024matters} as additional baselines; for the 13B parameter size, we selected LLaVA-Next-13B~\citep{liu2024llavanext} as an additional baseline model. Notably, as shown in Table~\ref{tab:tuning}, each of these additional baseline models was finetuned on a substantially larger dataset than ours.
We evaluate all models under the same evaluation protocol to ensure fair comparisons.

\highlight{Benchmark datasets.}
We evaluated on the BLINK, MMBench, Seedbench, TaskMeAnything, and MMMU datasets. For consistency, we employed both \textbf{in-domain templates} and \textbf{out-of-domain} templates in Sec.~\ref{sec:exp-setup} as evaluation templates. To further measure the ease of use of the template-tuned models, we selected three most commonly-used \textbf{simple templates} in VQA tasks: (1) \textit{\{question\}\textbackslash{}n\{choices\}}, (2) \textit{Question:~\{question\}\textbackslash{}nChoices:~\{choices\}}, and (3) \textit{Question:~\{question\}\textbackslash{}nSelect from the following choices:~\{choices\}}.  

\highlight{Evaluation Protocol.}
In this section, our evaluation settings are consistent with those in Sec.~\ref{sec:exp-setup}. For the evaluation metric, in addition to the answer accuracy, we follow~\citep{sclar2023quantifying} and report the range (Max-Min) between the best and worst accuracy across all evaluation instruction templates to quantify MLM's performance fluctuation to instruction template variations.

\begin{table*}[t]
\centering
\renewcommand{\arraystretch}{1.5}
\resizebox{\linewidth}{!}{%
\begin{tabular}{lcccccccccccccccccc}
\toprule
\multirow{2}{*}{\textbf{Model}} & \multirow{2}{*}{\textbf{\# IT-Data}}       &          & \multicolumn{3}{c}{\textbf{BLINK}} & \multicolumn{3}{c}{\textbf{MMB}} & \multicolumn{3}{c}{\textbf{SeedB}} & \multicolumn{3}{c}{\textbf{TMA}} & \multicolumn{3}{c}{\textbf{MMMU}} & \multirow{2}{*}{\textbf{Overall}} \\ \cmidrule(lr){4-6}\cmidrule(lr){7-9}\cmidrule(lr){10-12}\cmidrule(lr){13-15}\cmidrule(lr){16-18}
                                      & &          & S     & ID        & OOD       & S      & ID         & OOD       & S       & ID         & OOD        & S     & ID       & OOD      & S     & ID        & OOD

\\ \hline
\multicolumn{19}{c}{\cellcolor[HTML]{BFBFBF}\textbf{7B / 8B Models}}
\\ \hline
&       & Avg.     & 43.67      & 37.26     & 38.72     & \secbest{70.00}       & 68.55      & 69.20     & 60.67        & 57.35      & 56.16      & 37.00      & 42.94    & 42.60    & \secbest{36.67}      & \secbest{37.19}     & 36.16 &48.94    \\ \cline{3-3}\multirow{-2}{*}{\cellcolor[HTML]{FFFFFF}LLaVA-1.5-7B} & \multirow{-2}{*}{\cellcolor[HTML]{FFFFFF}665K} & Max-Min & 8.00       & 15.00     & 15.00     & 18.00       & 16.00      & 9.00      & 5.00         & 18.00      & 16.00      & 14.00      & 26.00    & 18.00    & 4.00       & 14.00     & 13.00 &13.93    \\
\hline &          & Avg.     & \secbest{45.33}      & 38.92     & 37.64     & 62.67       & 60.43      & 58.08     & \best{70.00}        & \best{65.29}      & \secbest{62.16}      & \secbest{50.67}      & 44.06    & 44.60    & 33.67      & 31.51     & 29.24 &48.95   \\ \cline{3-3}\multirow{-2}{*}{\cellcolor[HTML]{FFFFFF}LLaVA-Next-7B} & \multirow{-2}{*}{\cellcolor[HTML]{FFFFFF}760k} & Max-Min & 7.00       & 16.00     & 12.00     & 10.00       & 20.00      & 9.00      & 2.00         & 18.00      & 10.00      & 16.00      & 17.00    & 11.00    & 2.00       & 18.00     & 8.00 &11.73    \\
\hline
 &               & Avg.     & 36.00      & 34.44     & 34.04     & 50.07       & 47.51      & 47.16     & 30.67        & 29.66      & 28.80      & 31.67      & 29.76    & 30.76    & 25.67      & 28.06     & 28.40 &34.18    \\ \cline{3-3}\multirow{-2}{*}{\cellcolor[HTML]{FFFFFF}Qwen-VL-7B}
                                      & \multirow{-2}{*}{\cellcolor[HTML]{FFFFFF}50M} & Max-Min & 4.00       & 9.00      & 8.00      & 3.00        & 11.00      & 11.00     & 10.00        & 17.00      & 12.00      & 9.00       & 19.00    & 14.00    & 2.00       & 17.09     & 11.00 & 10.47    \\\hline
 &      & Avg.     & 31.67      & 40.09     & 40.28     & 62.67       & \best{74.02}      & \best{75.16}     & 56.00        & 58.77      & 58.32      & 39.33      & \secbest{51.55}    & \secbest{51.48}    & 39.00      & 36.49     & \secbest{36.36} &\secbest{50.08}   \\ \cline{3-3}\multirow{-2}{*}{\cellcolor[HTML]{FFFFFF}Qwen-VL-Chat-7B}
                                      & \multirow{-2}{*}{\cellcolor[HTML]{FFFFFF}50M} & Max-Min & 4.00       & 21.00     & 20.00     & 3.00        & 17.00      & 14.00     & 2.00         & 20.00      & 13.00      & 8.00       & 17.00    & 12.00    & 10.00      & 16.00     & 10.00 &12.47   \\\hline
  &        & Avg.     & 39.33      & \best{45.97}     & \best{46.36}     & \best{71.00}       & 70.73      & 70.28     & 43.33        & 53.36      & 54.04      & 36.00      & 47.40    & 46.20    & 29.33      & 27.48     & 28.36 &47.28   \\ \cline{3-3}\multirow{-2}{*}{\cellcolor[HTML]{FFFFFF}IDEFICS2-8B}
                                      & \multirow{-2}{*}{\cellcolor[HTML]{FFFFFF}1.8M} & Max-Min & 4.00       & 17.00     & 10.00     & 6.00        & 11.00      & 9.00      & 7.00         & 16.00      & 17.00      & 8.00       & 20.00    & 17.00    & 3.00       & 14.00     & 11.00 &11.33    \\
\hline\rowcolor[HTML]{E6E6E6}
 &       & Avg.     & \best{46.33}      & \secbest{43.19}     & \secbest{45.44}     & 68.67       & \secbest{71.66}      & \secbest{73.20}     & \secbest{64.33}        & \secbest{65.13}      & \best{64.16}      & \best{52.00}      & \best{51.78}    & \best{52.64}    & \best{39.33}      & \best{37.46}    & \best{37.32} &\best{54.18}    \\\cline{3-3}\rowcolor[HTML]{E6E6E6}\multirow{-2}{*}{LLaVA-1.5-7B w/ 5K templates}
                                      & \multirow{-2}{*}{665K} & Max-Min & 5.00       & 13.00     & 2.55      & 10.00       & 12.00      & 8.00      & 3.00         & 11.00      & 6.00       & 4.00       & 22.00    & 10.00    & 9.00       & 11.00     & 6.00 & 8.84    \\ \hline
\multicolumn{19}{c}{\cellcolor[HTML]{BFBFBF}\textbf{13B Models}}
\\
\hline
&       & Avg.     & \best{40.00}      & 38.75     & \secbest{41.20}     & \best{72.33}       &\secbest{73.42}      & \secbest{71.24}     & 67.00        & \secbest{68.87}      & \secbest{66.92}      & \secbest{54.00}      & \best{52.38}    &  \best{52.24}   & \secbest{37.33} & \secbest{39.00}    &  \secbest{37.20} &\secbest{54.13}  \\ \cline{3-3}\multirow{-2}{*}{\cellcolor[HTML]{FFFFFF}LLaVA-1.5-13B} & \multirow{-2}{*}{\cellcolor[HTML]{FFFFFF}665K} & Max-Min & 7.00       & 16.00     & 14.00     & 3.00       & 12.00      & 6.00     & 5.00         & 9.00      & 10.00      & 8.00      & 16.00    & 15.00    & 6.00       & 16.00     & 10.00 & 10.20   \\\hline
 &          & Avg.     & \secbest{39.67}      & \secbest{40.72}     & 38.16    & 64.67       & 63.47      & 63.40     & \secbest{68.33}        & 68.76      & 66.88     & \best{54.67}      & \secbest{51.53}   & 47.68    & 31.00      & 33.23     & 33.80 &51.06   \\ \cline{3-3}\multirow{-2}{*}{LLaVA-Next-13B} & \multirow{-2}{*}{\cellcolor[HTML]{FFFFFF}760k} & Max-Min & 1.00       & 15.00     & 13.00     & 9.00       & 19.00      & 15.00      & 1.00         & 12.00      & 11.00      & 5.00      & 21.00    & 14.00    & 2.00       & 21.00     & 10.00 &11.27       \\ \hline\rowcolor[HTML]{E6E6E6}\cellcolor[HTML]{E6E6E6} &       & Avg.     & 37.67      & \best{41.22}     & \best{42.68}     & \secbest{70.00}       & \best{73.88}      & \best{74.68}     & \best{69.33}        & \best{69.37}      & \best{69.48}      & 51.33      & 50.49    & \secbest{50.68}    & \best{39.67}      & \best{43.21}    & \best{44.40} &\best{55.21}    \\\cline{3-3}
 \rowcolor[HTML]{E6E6E6} \multirow{-2}{*}{LLaVA-1.5-13B w/ 100 templates}
 & \multirow{-2}{*}{665K} & Max-Min & 14.00       & 15.00     & 8.00      & 12.00       & 10.00      & 10.00      & 3.00         & 7.00      & 5.00       & 1.00       & 12.00    & 5.00    & 7.00       & 15.00     & 15.00  & 9.27   \\ \bottomrule
\end{tabular}%
}
\caption{Comparison of our tuned models trained under the optimal template scale against similar-scale MLMs. \textbf{Avg.} denotes the average accuracy and \textbf{Max-Min} denotes the difference between best and worst accuracy across all templates. \textbf{\# IT-Data} is the size of instruction tuning data the model used. \textbf{S} indicates the evaluation of three commonly used simple templates, \textbf{ID} refers to the evaluation of 100 instruction templates that our template-tuned model has encountered during training, and \textbf{OOD} denotes the evaluation of 25 manually crafted templates not included in our instruction template generator's template space. The best results are marked in \best{red bold} and the second best in \secbest{blue}. \textbf{Training with optimal template scale can boost performance across most benchmarks.}}
\label{tab:tuning}
\end{table*}

\subsection{Main Results}
\label{sec:vsft-main-exp}

As presented in Table~\ref{tab:tuning}, we compare the performance of our tuned 7B and 13B models, which we trained with the optimal template scale, against several prominent MLMs of similar scale, revealing the following two key findings.

\highlight{Training on the optimal template scale significantly enhances MLM's performance without increasing the scale of training data.} 
Compared to LLaVA-1.5-7B and LLaVA-1.5-13B, which utilize the same pretrained models as our template-tuned models but rely on original instruction tuning data, training with the optimal template scale achieves substantial performance improvements across most datasets in all three evaluation settings. Additionally, our tuned models trained with the optimal template scale outperforms other prominent MLMs of similar scale, despite these models being trained on significantly larger datasets (up to 75.19 times larger). This underscores the efficiency and effectiveness of our approach of training MLMs with the optimal template scale to achieve superior performance without the need for extensive data scaling. By focusing on the quality and diversity of instruction templates rather than the quantity of training data, our method demonstrates a more resource-efficient pathway to enhancing visual instruction tuning. 

\highlight{Training on the optimal template scale significantly mitigates MLM's sensitivity to diverse instruction templates.} 
Compared to LLaVA-1.5-7B and LLaVA-1.5-13B, which rely on original instruction tuning data, our approach of training MLMs under the optimal template scale not only achieves superior overall performance but also significantly reduces the performance fluctuation range (Max-Min) across multiple evaluation instruction templates in most cases. This reduction in fluctuation range indicates that training on the optimal template scale enhances model stability and adaptability when faced with varying instruction formats, a critical requirement for real-world applications where input instructions can vary widely. 
Furthermore, when compared to other prominent MLMs of similar scale, our tuned models trained with the optimal template scale consistently exhibit a lower performance fluctuation range. This consistency holds true across both in-domain (\textbf{ID}) and out-of-domain (\textbf{OOD}) instruction template settings, demonstrating the robustness of our approach across diverse evaluation scenarios. However, counterexamples are more likely to arise with commonly used simple templates (\textbf{S}), likely due to the limited diversity of only three evaluation templates.
Notably, even when evaluated using manually crafted out-of-domain templates—which lie entirely outside the template space of our instruction template generator, our template-tuned models frequently demonstrate a smaller performance fluctuation range. This observation underscores the ability of training on the optimal template scale to generalize beyond the specific instruction templates encountered during training, rather than merely memorizing them.

\section{Related Work}
\label{sec:related}

\highlight{Multimodal language model.}
In recent years, multimodal language models (MLMs) have advanced visual-language learning by integrating visual encoders within various pretrained large language models~\citep{sun2023finegrained, lyu2023macawllm, tang2023llmvagebc, wang2023chatvideo, bi2023misar, chen2023groundingprompter, liu2024prismer, peng2023kosmos2, chen2023pali3, shukor2023unival, lin2023mmvid, lu2023chameleon, li2023mimicit, sun2024emu, awadalla2023openflamingo, sun2024generative, Xue2024xGenMMA}. With the increasing availability of open-sourced LLM backbones and extensive visual instruction tuning data, models like the BLIP series~\citep{dai2024instructblip, li2022blip, li2023blip2, Panagopoulou2023XInstructBLIPAF, Xue2024xGenMMA}, QwenVL series~\citep{bai2023qwen, Qwen2VL}, LLaVA series~\citep{liu2024visual, liu2023improvedllava, liu2024llavanext}, and InternVL series~\citep{chen2023internvl, chen2024far}, have achieved unprecedented performance in a wide range of visual tasks~\citep{lin2024draw,luo2024llm, ma2024m,xue2024xgen, wang2024finetuned, an2024mc, zhang2023split}. These models, which take both visual content and language as input and output language, are now considered a new type of foundation model with exceptional visual understanding capabilities. However, these MLMs largely overlooked the significance of instruction templates of prompts, resulting in unreliable, unstable evaluation results.

\highlight{Influence of template perturbation.}
Recent research illustrated how prompt perturbations affect the performance and robustness of large language models (LLMs) and MLMs~\citep{gonen2022demystifying, lu2021fantastically, madaan2023makes, zhuo2024prosa, gan2023sensitivity}. \citet{liang2022holistic} performed a comprehensive examination of MLM outputs under diverse prompt designs, emphasizing the importance of systematic evaluation to ensure MLM robustness. \citet{liu2024seeing} highlight that MLMs often produce incorrect responses when presented with nuanced, leading questions, underlining their susceptibility to prompt design variations. To solve this problem, \citet{chatterjee2024posix} propose a prompt sensitivity index method that captures the relative change in log-likelihood of the given prompts, making it a more reliable measure of prompt sensitivity. Some former methods~\citep{leidinger2023language, mizrahi2024state, voronov2024mind} have also proposed to extend the evaluation benchmarks from a single prompt to multiple variants for each prompt. However, these former methods are all based on hand-crafted methods, which are not comprehensive enough to evaluate LLMs and MLMs. Meanwhile, most existing benchmarks, such as BLINK~\citep{fu2024blink}, SeedBench~\citep{li2023seed}, MMBench~\citep{liu2025mmbench}, TaskMeAnything~\citep{zhang2024task}, and MMMU~\citep{yue2024mmmu}, still keep using a single template of the prompts for the performance evaluation.

\section{Conclusion}
We introduced a programmatic instruction template generator to efficiently produce diverse, high-quality instruction templates at scale, aimed at investigating the scaling effect of instruction templates for MLM's visual instruction tuning.
Our investigation into scaling instruction templates for MLM training showed that MLM capabilities did not monotonically improve with increasing template scale and instead peaked at a medium template scale, which varies with the model's parameter size.
Additionally, using this instruction template generator, we proposed a simple yet effective method to improve visual instruction tuning by augmenting the original instruction tuning dataset at the optimal template scale, offering an efficient and cost-effective solution to improve MLMs.

\bibliography{colm2025_conference}
\bibliographystyle{colm2025_conference}

\appendix


\section{Details of Instruction Template Generator}
\label{sec:generator}

\begin{figure*}[t]
  \centering
    \includegraphics[width=\textwidth]{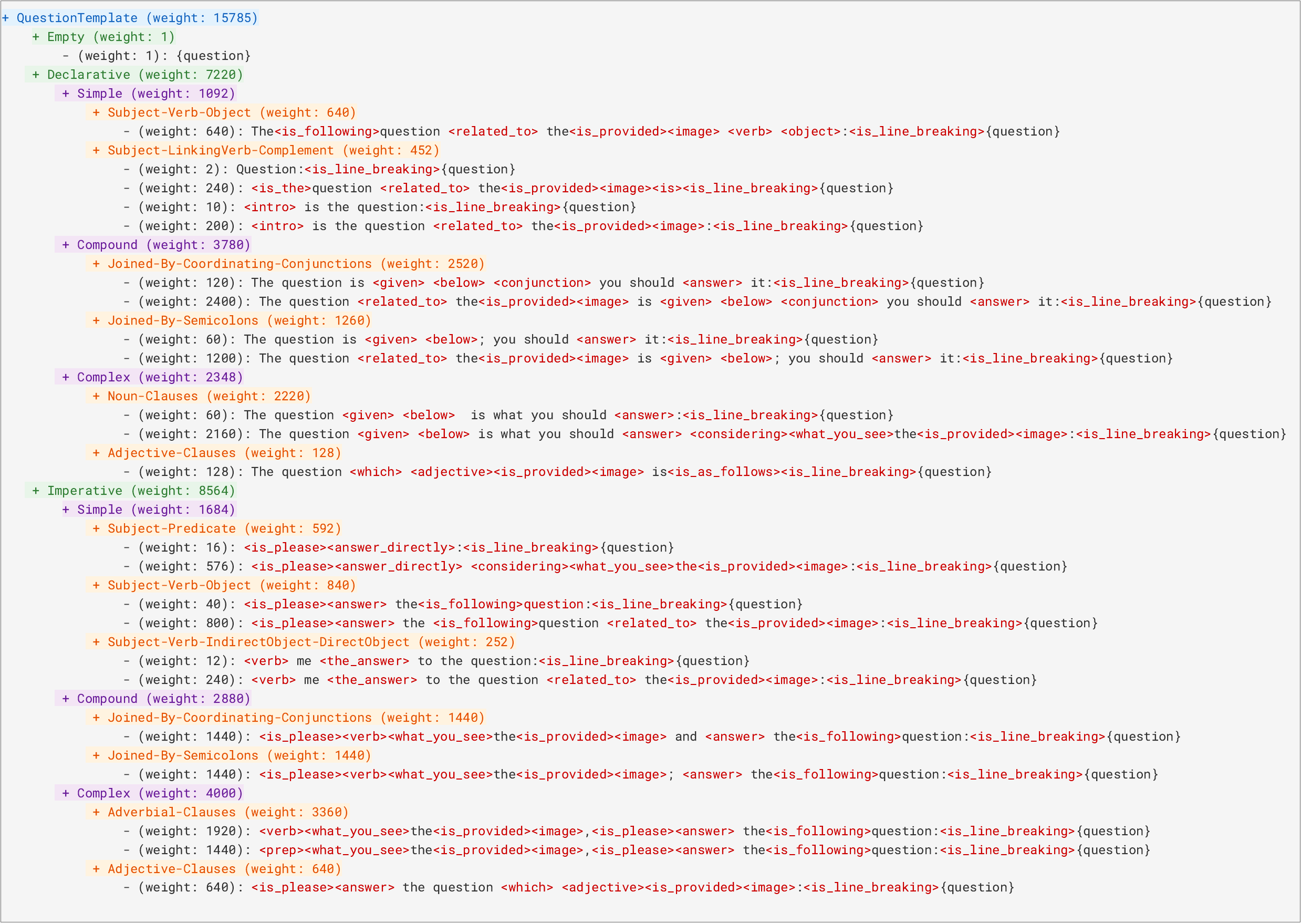}
  \caption{Sentence pattern trees with meta templates. Each tree uses distinct colors to denote different levels. Placeholders are marked in \textcolor{red}{red}, while static segments are marked in black. We further mark the weight of each node (\# generated templates).}
  \label{fig:sentence-pattern-trees}
\end{figure*}

Our instruction template generator can produce an extensive template space comprising 15K visual instruction templates. Our method operates by sampling meta templates from the sentence pattern tree with a weighted sampling algorithm and then programmatically populating placeholders in meta templates with randomly sampled positional synonyms.
In this section, we present the details of our sentence pattern trees with diverse meta templates.
We construct a sentence pattern trees for visual instructions, consisting of 24 meta templates.
We present the taxonomy and meta templates of the sentence pattern tree for visual instructions in Figure~\ref{fig:sentence-pattern-trees}.

\section{Additional Experiments on MLM’s Sensitivity to Instruction Templates}
\label{sec:more-exp}

\begin{figure}[ht]
  \centering
  \includegraphics[width=0.7\columnwidth]{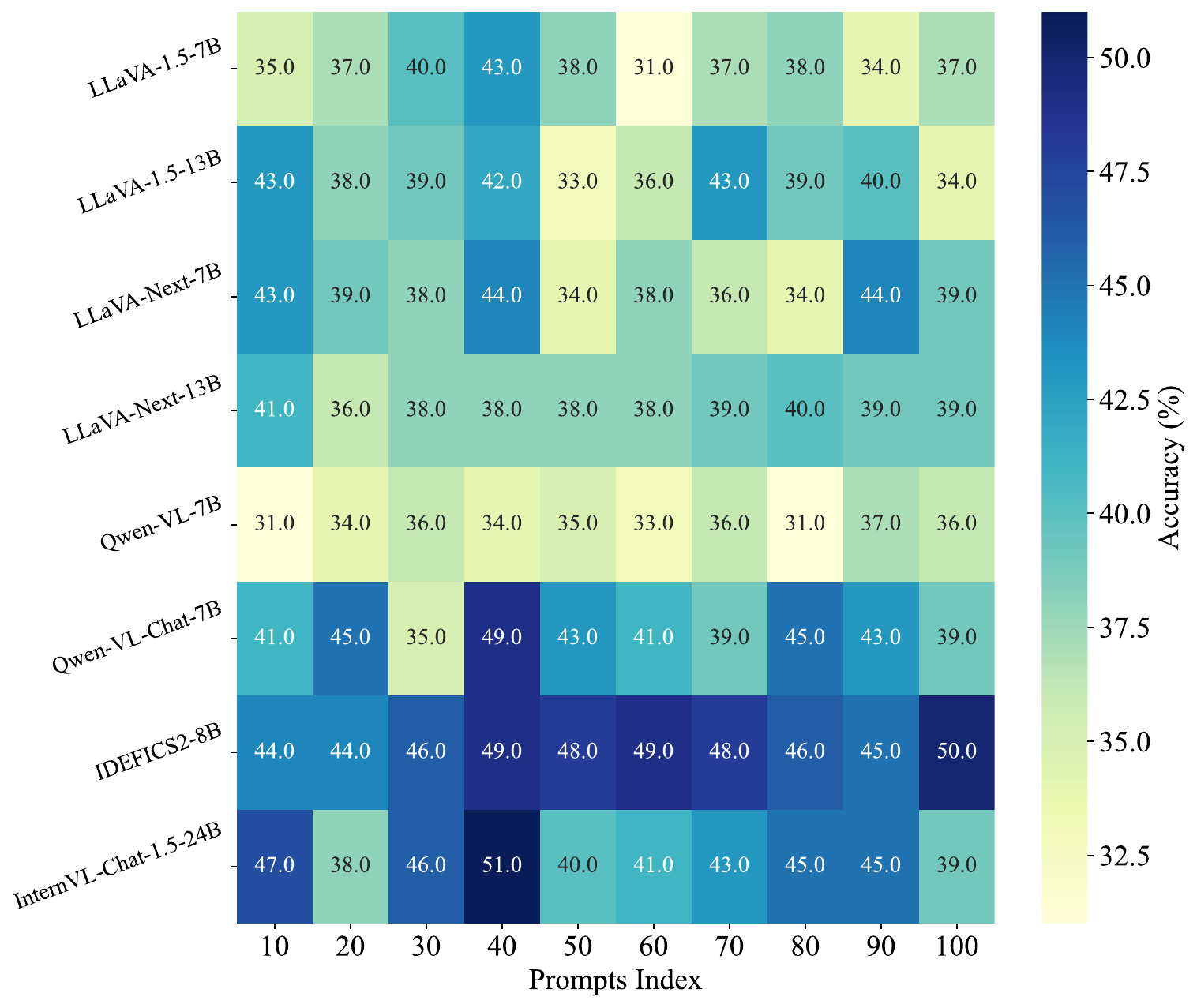}
  \caption{Heat map illustrating the performance variations of eight MLMs on the BLINK dataset across ten instruction templates (selected from the \textbf{in-domain} templates set in Sec.~\ref{sec:vsft-scaling}). The darker the color, the better the performance. \textbf{No single instruction template performs optimally for all MLMs.}}
  \label{fig:heatmap}
\end{figure}

In this section, we explore whether a universally effective instruction template for most MLMs exists. To this end, we analyze the performance of eight prominent MLMs on the BLINK dataset using ten instruction templates selected from the \textbf{in-domain} templates set as described in Sec.~\ref{sec:vsft-scaling}. A heat map is presented in Figure~\ref{fig:heatmap} to illustrate the performance variations, where darker shades correspond to superior performance.

The results reveal substantial performance variability across MLMs for the same instruction template, as indicated by the diverse color gradients within each column of the heat map. This variability highlights a critical observation: \textbf{no single instruction template consistently performs optimally for all MLMs}. This lack of universality implies that each model exhibits distinct sensitivities and preferences toward different instruction templates, which complicates the task of designing or selecting a universally effective template.

The observed variations have important implications for the instruction template design and evaluation of MLMs. Specifically, they underscore the limitations of a one-size-fits-all approach to instruction optimization. Efforts to identify an ideal instruction template through a single, static search are unlikely to yield universally effective results. Instead, tailored strategies that consider the specific characteristics and requirements of individual MLMs can be necessary to achieve optimal performance.

\end{document}